# FinanceQA: A Benchmark for Evaluating Financial Analysis Capabilities of Large Language Models

**Spencer Mateega, Carlos Georgescu, Danny Tang**

AfterQuery

`{spencer,carlos,danny}@afterquery.com`

## Abstract

FinanceQA is a testing suite that evaluates LLMs' performance on complex numerical financial analysis tasks that mirror real-world investment work. Despite recent advances, current LLMs fail to meet the strict accuracy requirements of financial institutions, with models failing approximately 60% of realistic tasks that mimic on-the-job analyses at hedge funds, private equity firms, investment banks, and other financial institutions. The primary challenges include hand-spreading metrics, adhering to standard accounting and corporate valuation conventions, and performing analysis under incomplete information - particularly in multi-step tasks requiring assumption generation. This performance gap highlights the disconnect between existing LLM capabilities and the demands of professional financial analysis that are inadequately tested by current testing architectures. Results show that higher-quality training data is needed to support such tasks, which we experiment with using OpenAI's fine-tuning API. FinanceQA is publicly released at https://huggingface.co/datasets/AfterQuery/FinanceQA.

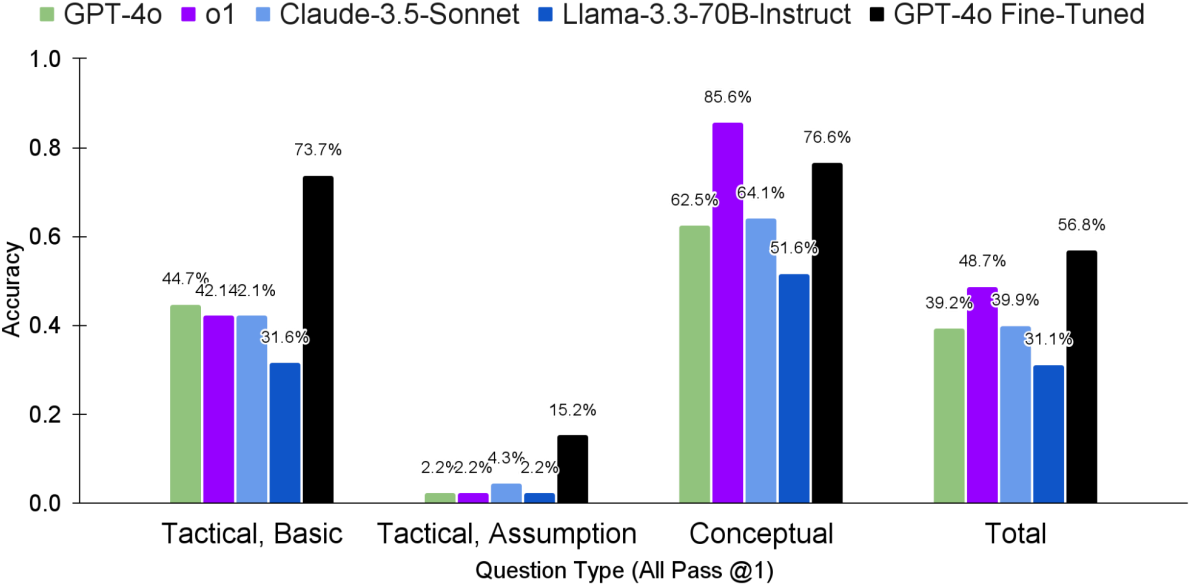

Figure 1: Performance of model comparison set and GPT-4o fine-tuned on FinanceQA

# 1 Introduction

Large Language Models (LLMs) are mostly pre-trained on public data available on the internet, which makes them perform increasingly well on prompts that such data can answer, but in turn, makes them have subpar performance on work-critical tasks that require experiential knowledge and adherence to industry-specific practices (Chen et al., 2024). People in specialized fields such as finance, healthcare, and law actively have difficulty incorporating LLMs and other AI tools in their workflows. Their outputs are repeatedly sprinkled with non-contextual and non-compliant information, answers that don't follow industry best practices or worse - entirely wrong outputs. These fields require a high degree of precision and have little tolerance for errors, making the deployment of LLMs more of a business risk than an opportunity. In this paper, we propose a new way to evaluate these models with actual on-the-job tasks that professionals do daily, first focusing on the financial domain.

# 2 Related Work

There have been numerous efforts in fine-tuning LLMs on specialized data, as witnessed by Guha et al. (2023) and Islam et al. (2023). While these efforts are extensive and valuable, we hypothesize that many existing datasets may not fully align with the tasks performed by professionals in real-world scenarios. To test this hypothesis, we start with the finance domain. As an example, we reviewed FinanceBench (Islam et al., 2023), which was found to primarily focus on basic information extraction tasks that, while important, could be accomplished through simple text searches. For example, their dataset includes questions like "What is the FY2018 capital expenditure amount (in USD millions) for 3M?" Their benchmark also contains relatively simple numerical calculations that don't require accounting knowledge - for example, their dataset includes basic questions about balance sheet debt but overlooks more realistic calculations, such as factoring in operating lease liabilities into total debt obligations. Further, the dataset contains qualitative questions with subjective answers, such as "What was the key agenda of the AMCOR's 8k filing dated 1st July 2022?"

Many professional-services-focused datasets follow this pattern of not having difficult and realistic question answer pairs, and as a result, fine-tuning models on such question-answer pairs makes the model suspect to the same dilemma that the base model is suspected to: not being able to answer day-to-day job-related queries.

# 3 Issues Applying LLMs to the Financial Domain

The deployment of LLMs in real-world finance settings faces a unique set of challenges. Current LLMs fall short across multiple dimensions: they lack the precision required for professional application, fail to capture the complex recalculation workflows used by analysts, don't enforce accounting and corporate finance standards, and inadequately represent how analysts reason with incomplete information (see Section 5.2 Performance).

## 3.1 High Required Accuracy Threshold

The required accuracy threshold for meaningful adoption is exceptionally high due to the nature of investment work. Investment analysis demands near-perfect precision for several reasons:

First, investment decisions involve significant capital allocation, where errors can result in material financial losses. As a result, unlike other domains where the relationships between accuracy and usefulness might be linear, finance exhibits a sharp discontinuity, similar to the materiality thresholds established in financial reporting (SAB 99, 1999). This principle, where errors below 5% may be considered immaterial, but those exceeding this threshold require intense supervision, parallels our findings with LLMs.

Second, while impressive in many contexts, an LLM that achieves 80% accuracy provides minimal practical value since its output must still be meticulously verified by a human. The time it takes to double check work often requires more time than if the analysis was performed from scratch. This is because most data verification occurs line-by-line, meaning that the analyst has to go through every single number, verify its truth, whether or not it should be included in the calculation, and check if there are any missing numbers that should be included. On the bright side, however, once an LLM crosses a higher accuracy threshold, its utility increases dramatically as it begins to earn the trust of investment professionals and can be used to replace existing work.

Finally, the workflow in professional investment firms is hierarchical — mid-level and senior professionals rely on "analysts" (hedge fund analysts and senior analysts; private equity associates and senior associates; investment banking analysts and associates, etc.) to provide defensible analysis that forms the foundation for investment decisions. Since seniors generally are not calculating any metrics or producing work product, the tolerance for errors for analysts in this context is very little.

## 3.2 High Level of Financial Rigor

Professional investment analysis requires a level of financial rigor that goes beyond simply pulling data from industry data providers or taking management (the company providing their financials) calculations at face value. Often, investment firms must independently verify and recalculate metrics through "hand-spreading" — calculating metrics by directly analyzing data from primary sources such as SEC filings or privately provided materials (e.g., a CIM provided by a banker). Current LLMs fall short as they often do not perform these essential recalculations (Table 1).

Consider the calculation of diluted shares outstanding. While financial data providers (e.g., Yahoo Finance and S&P Capital IQ) and management report this metric (Figure 2), it is typically only presented as a final number without supporting calculation methodology.

| Earnings per share: | | | | | | |
|---|---|---|---|---|---|---|
| Basic | $ | 11.86 | $ | 9.72 | $ | 9.70 |
| Diluted | $ | 11.80 | $ | 9.68 | $ | 9.65 |
| Weighted average shares outstanding: | | | | | | |
| Basic | | 7,431 | | 7,446 | | 7,496 |
| Diluted | | 7,469 | | 7,472 | | 7,540 |

Figure 2: Microsoft's 2024 10-K[1] stating diluted shares outstanding without showing supporting calculations

This "black box" is problematic and functionally unusable for professionals who require full transparency and verifiability in their analysis. As a result, analysts must manually reconstruct these calculations through a review of primary documents. The process of calculating diluted shares outstanding illustrates this complexity. It requires: 1) identifying all potential sources of dilution from various sections of a 10-K (such as Restricted Stock Units (RSUs), Performance Stock Units (PSUs), convertible bonds, warrants, and stock options), 2) analyzing the specific terms of each instrument, 3) determining which instruments are dilutive under current market conditions, and 4) performing the actual calculations according to accounting standards (Koller, et al., 2020).

While this process is currently labor-intensive, it is fundamentally algorithmic in nature. An LLM, properly trained in high-quality financial reasoning data, could automate this process while maintaining the standards required by investment professionals. However, our results imply current LLMs lack the reasoning data needed to perform these calculations (Table 1).

---

[1] https://www.sec.gov/Archives/edgar/data/789019/000095017024087843/msft-20240630.htm

## 3.3 Adherence to Accounting Conventions

Professional financial analysis requires strict adherence to established accounting and corporate finance conventions that current LLMs frequently overlook (see section 5.2 Performance). This isn't merely about getting "roughly correct" answers - there are well-defined procedures for calculating financial metrics that must be followed (Koller, et al., 2020).

Consider the calculation of Accounts Payable Days. While a naive approach might simply be to use end-of-period Accounts Payable, the correct methodology, according to accounting principles, requires using the average Accounts Payable balance over the period. Averaging is correct because it represents the true level of payables throughout the year (Holthausen et al., 2014), reducing the impact of end-of-period anomalies and maintaining consistency between the numerator (a stock measure) and denominator (a flow measure).

Current LLMs frequently fail to make these distinctions (see section 5.2 Performance), often defaulting to more straightforward but incorrect calculations. To address this limitation, our benchmark, FinanceQA, draws from authoritative sources, including leading corporate finance textbooks (Holthausen et al., 2014; Koller et al., 2020), to ensure training data reflects the rigorous standards applied in professional practice.

### 3.4 Handing Incomplete Information

Financial analysis frequently requires working with incomplete information, a challenge that current LLMs and evaluation benchmarks fail to address (as will be seen in Figure 3). Analysts routinely face scenarios where direct data isn't available, requiring them to use estimation methods and assumptions to reach their conclusions. While second nature to trained financial professionals, these approaches require specific domain expertise that current LLMs appear to lack.

The disconnect between existing LLM capabilities and real-world requirements is particularly evident in current financial benchmarks. Many evaluation frameworks, such as FinQA ([Chen et al., 2021](#)), employ a "context-first" approach - providing annotators with financial documents and asking them to generate questions based on stated information. This methodology is fundamentally misaligned with how analysts work, which is the opposite. Instead of looking at documents and then generating questions, analysts 1) begin with critical questions about a company, 2) search through available information to answer these questions, and 3) apply structured estimation techniques when direct information isn't provided ([CFA Institute, 2024](#)). This misalignment in evaluation methodology leads to artificially inflated performance metrics that don't reflect LLMs' true capabilities in real-world tasks. Instead, we seek to provide a benchmark corresponding to the questions analysts must answer on the job, and then we evaluate how well LLMs can leverage available information - complete or incomplete - to reach reasonable conclusions.

# 4 FinanceQA: A financial reasoning benchmark

To address the limitations of current LLMs in financial applications, we propose a new benchmark that emphasizes both technical precision and analytical reasoning. Our approach is built on two complementary pillars:

First, we've developed a suite of tactical questions based on primary financial documents (e.g., 10-Ks) that test an LLM's ability to 1) perform precision recalculations from raw financial data, 2) apply correct accounting and corporate finance conventions, 3) make appropriate assumptions when faced with incomplete information, and 4) follow real-world calculation standards.

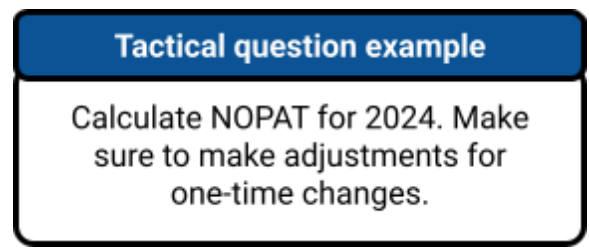

The tactical questions are further segmented into "basic" and "assumption-based" questions. As the name implies, the assumption-based questions require an assumption not provided in the context. For example, the question "Let's estimate [Costco's] Variable Lease Assets in 2023." is assumption-based as Costco's 10-K includes a Lease section with variable lease costs but not variable lease assets[2]. Hence, an analyst would make a logical assumption—operating lease costs and operating lease assets are provided, and they would assume that the ratio of variable lease assets to operating lease assets equals the ratio of variable lease costs to operating lease costs and note that assumption.

---

[2] https://www.sec.gov/Archives/edgar/data/909832/000090983224000049/cost-20240901.htm

Each corresponding tactical question has a context window for the 10-K information. To reduce the impact of potential hallucinations from having a large context window and test more for reasoning and finance expertise, we only included relevant components of the 10-K in our context window, including the consolidated financial statements, notes, management discussion and analysis section, etc.

Second, we've created a collection of conceptual questions that evaluate an LLM's capacity for financial reasoning. These questions require models to 1) understand relationships between financial metrics, 2) apply logic to derive implied values, 3) make reasonable estimations using industry knowledge, and 4) demonstrate an understanding of accounting and valuation principles.

**Conceptual question example**

A company has a 2x sales multiple and a 5x EBITDA multiple. What is its EBITDA margin?

Using high-quality financial reasoning data, establishing a high ceiling for model capabilities is essential.

By combining these tactical and conceptual elements, our benchmark provides the first comprehensive assessment of an LLM's ability to truly replicate the work of junior investment professionals.

### 4.1 Dataset Annotation

Human annotators were tasked to create the dataset. All annotators had experience at one or more reputable hedge funds, private equity firms, or investment banks and had taken accounting and finance courses from a leading business school. They were instructed to create questions that they would actually find useful for their real-world work, which we call tactical questions. When creating tactical questions, annotators were provided with a public company name and a corresponding primary document (e.g., a 10-K). Annotators were also provided corporate finance textbooks (Holthausen et al., 2014; Koller et al., 2020) for reference and were asked to label sections of the 10-K that were relevant, which were then aggregated to create the context window for tactical questions.

For conceptual questions, annotators were asked to create questions and answers similar to conceptual questions asked at their firms, ranging from easy to hard. All questions were created without the external aid of LLMs and without copying any LLM written work. Every prompt and answer was manually verified to ensure they were unambiguous and had the correct solution. All annotators were based in the U.S. and were paid above market rates.

## 5 Results

### 5.1 Setup

We study 4 commercial API-access models: OpenAI's GPT-4o and o1, Anthropic's

Claude-3.5-Sonnet, and Meta's Llama-3.3-70B-Instruct. For all models, we had a user prompt with the content "Context:\n{context}\n\nQuestion: {question}\n \n Provide a concise answer." For GPT and Llama, we added a system prompt with the content "You are a helpful assistant. Provide concise answers." For the tactical questions, the context parameter was filled with the subset of a specific primary document that contained the supporting text and tables needed to answer all questions about the company in question. This includes all financial statements and relevant document notes. For the conceptual questions, the context parameter was left blank.

Tasks were evaluated using exact correct matches, meaning that there were no partial points given. The number 1 was given when answers correctly factored in the necessary assumptions and lines of reasoning an analyst would make on a job and a corresponding 0 for an incorrect match.

## 5.2 Performance

Initial Results: Table 1 shows FinanceQA's performance on a range of models. Particularly, the top SOTA model, o1, only performed at a ~48.7% accuracy rate on FinanceQA questions, with all the models having increased difficulty in the assumption questions that required handling incomplete information. In Figure 3, we can clearly see that all models performed relatively well on conceptual questions, indicating that the models were possibly trained more on such questions but trained less on basic and assumption-based FinanceQA questions that more closely resemble on-the-job analysis of finance professionals, where it performed worse.

| LLM | Tactical, Basic | Tactical, Assumption | Conceptual | Total |
|---|---|---|---|---|
| GPT-4o | **0.447** | 0.022 | 0.625 | 0.392 |
| o1 | 0.421 | 0.022 | **0.856** | **0.487** |
| Claude-3-5-sonnet-20241022 | 0.421 | **0.043** | 0.641 | 0.399 |
| Llama-3.3-70B-Instruct | 0.316 | 0.022 | 0.516 | 0.311 |

Table 1: Performance on FinanceQA

Figure 3: Model accuracy across different financial question types

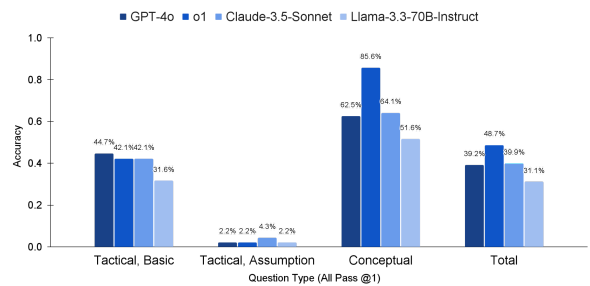

Figure 4: Correlation matrix showing weak relationships between question types

As per the correlation matrix in Figure 4, we can see that there is no strong correlation between question types; these varying correlations validate the fact that the FinanceQA benchmark effectively evaluates diverse aspects of financial reasoning and fundamentally tests different capabilities.

## 5.3 Performance Explanation

As explained in Section 3, there are a few reasons why LLMs struggle with domain-specific tasks, specifically with tactical or on-the-job tasks.

First, regarding the high required accuracy threshold, LLMs must find precise measurements to make specific adjustments. For example, one common question that all of the LLMs failed to answer was, "What is the adjusted EBITDA during the year ending 2024?" Since EBITDA is a non-GAAP measure, it is typically up to the analyst to develop sound reasoning to adjust EBITDA to reflect cash flows most accurately. However, in our tests, the LLMs miss one or two adjustments, and a single missed adjustment then has the potential to cause a significant change to the final number.

This ties into our second explanation of the high financial rigor required by LLMs. EBITDA and EBIT, like many other metrics, are often given by the company itself. The company is incentivized to inflate these numbers to present their company as one that is generating more cash flow (S&P Global Ratings, 2024). Analysts can spend significant time deconstructing these figures and deciding what their interpretation of EBITDA is to form an opinion on valuation. However, the reasoning behind these interpretations is lacking in LLM responses, where the response usually follows a strict formula, such as EBITDA = EBIT + Depreciation & Amortization Expense + One Time Expenses, without understanding why certain adjustments should be made. For example, under newer accounting standards, operating leases are now capitalized on the balance sheet instead of remaining an operating expense, which means that they should be treated as debt or debt equivalents (Ernst & Young, 2021). In the context of EBITDA, the LLM needs to understand that one purpose of the metric is to remove the influence of debt, thus adding back operating lease costs.

This leads to the next reason why LLMs underperformed in our evaluation: not adhering to accounting conventions. Accounting rules are constantly changing, and while LLMs generally understand the rules when prompted, they struggle to make connections to numerical calculations. For example, while the LLM understands that an operating lease expense is split up into an interest payment and a depreciation of the right-of-use operating lease asset on the balance sheet, unless directly prompted, it won't follow know that it needs to calculate those metrics to add back to EBITDA, despite them being classified as interest and depreciation components under the latest accounting rules.

Finally, and perhaps most importantly, is an LLM's ability to handle incomplete information, which is a requisite skill for many finance and other professional services careers. As shown in Table 1, LLMs currently answer less than 5% of questions correctly when incomplete information exists and assumptions must be made. In our dataset, a relevant example is the concept of variable leases. Under accounting frameworks such as

IFRS 16[3] and ASC 842[4], certain variable lease costs are recorded as expenses, but corresponding lease assets and liabilities for these costs may not always be explicitly stated in financial disclosures. The job of a financial analyst would then be to build an assumption to either find those assets or liabilities in the "other" category or calculate the asset or liability itself. The ability to spot missing assumptions, generate those missing assumptions, and come up with a final conclusion is key to many investing or finance roles.

On the other hand, conceptual questions perform better, considering that they are similar to the math, logic, and coding questions that LLMs are more familiar with. This is remarkably more apparent with o1, which performs extremely well on conceptual questions.

## 6 Fine Tuning

### 6.1 Dataset Annotation

We sought to validate our hypothesis that fine-tuning on high-quality financial reasoning data can improve model performance on real-world tasks. We asked our annotators to continue to generate questions representative of day-to-day finance calculations. These new questions were designed to be independent of the evaluation set, including different industries and problems while maintaining the same level of rigor and real-world applicability as FinanceQA. Each question was manually reviewed by our annotators to ensure they matched the same standards that we had set for FinanceQA. To create a diverse and representative fine-tuning dataset, we synthetically generated multiple variations for each human-generated data row. This included variations for each question, context, and answer to ensure that each type of question had multiple versions with different potential financial statement line items, assumptions, numerical structures, and reasoning pathways. We utilized synthetic data due to having a limited budget, as we still wanted to cover different cases and prevent overfitting from reiterating on the same fine-tuning pairs. Our ultimate fine-tuning dataset consisted of 9,078 rows.

| Real Data Answer | Synthetic Data Answer |
| --- | --- |
| Merchandise inventories: $5,112.8M, Other current assets: $335.0M, Cash: $684.9M Calculation: Working cash = 2% of $30,603.8M = $612.1M (used since it's less than total cash). Operating assets: $5,112.8M + $335.0M + $612.1M = $6,059.9M Operating liabilities: $3,183.7M Final OWC Calculation: $6,059.9M - $3,183.7M = $2,876.2M | Merchandise inventories: $5,312.6M, Other current assets: $385.0M, Prepaid expenses: $150.0M, Cash: $702.3M Calculation: Working cash = 2% of $31,103.8M = $622.1M (used since it's less than total cash). Operating assets: $5,312.6M + $385.0M + $150.0M + $622.1M = $6,469.7M Operating liabilities: $3,345.2M Final OWC Calculation: $6,469.7M - $3,345.2M = $3,124.5M |

Figure 5: Comparison of real vs. synthetic data answer from Dollar Tree's 2024 10-K. The table highlights how synthetic data preserves reasoning while varying numerical values and assumptions, like new line items.

---

[3] https://www.ifrs.org/issued-standards/list-of-standards/ifrs-16-leases/

[4] https://www.fasb.org/page/PageContent?pageId=/projects/recentlycompleted/leases-topic-842lessorsleases-with-variable-lease-payments--postissuance-summary.html

## 6.2 Evaluation and Results

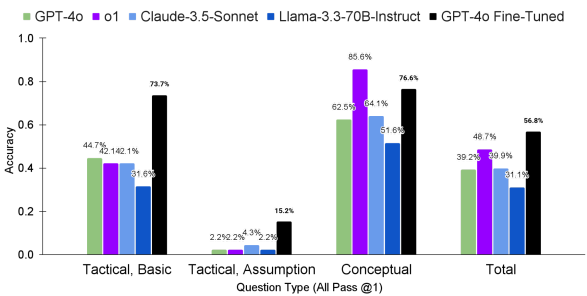

Figure 6: Performance of model comparison set and fine-tuned GPT-4o on FinanceQA

We fine-tuned the base GPT-4o model on our training set (Section 6.1), with default OpenAI fine-tuning API metrics. The evaluation dataset was FinanceQA, the same dataset used to initially evaluate the base models in Section 5. Our annotators hand-evaluated the responses, marking binary responses for correct and incorrect answers. The answers were judged for an exact correct match, the same as we did in Section 5.

| Type | 4o | 4o-Fine-Tuned | % Difference |
|---|---|---|---|
| Basic | 44.7% | 73.7% | **+ 68.9%** |
| Assumption | 2.2% | 15.2% | **+ 690.9%** |
| Conceptual | 62.5% | 76.6% | **+ 22.56%** |
| Total | 39.2% | 56.8% | **+ 44.9%** |

Figure 7: Percent difference between 4o base model and 4o base model fine-tuned on FinanceQA

As seen in Figure 7, the fine-tuned model performed noticeably better on each question type. In particular, the assumption-based questions saw over 600% improvement, likely due to the model remembering to infer from missing data and construct logical assumptions aligned with real-world financial analysis. Similarly, the model showed a 68.9% improvement in basic tactical questions and a 22.56% improvement in conceptual reasoning, suggesting that fine-tuning led to more structured financial reasoning across all task types.

While the magnitude of improvement was significant, it aligned with our expectations. The necessary adjustments to achieve accurate results are straightforward and can be systematically incorporated into the training process. This led to substantial performance gains, as the model had previously failed to correctly answer even basic questions. These issues can be addressed through proper training data, such as implementing the stock-flow averaging method discussed in Section 3.3.

While the improvements observed in Figure 7 are substantial, we performed a statistical significance test to determine where these improvements were meaningful beyond random variation.

The standard deviation of the paired differences is computed as follows:

$$s_d = \sqrt{\frac{\sum(d_i - \bar{d})^2}{n-1}}$$

The paired t-statistic is computed as:

$$t = \frac{\bar{d}}{s_d/\sqrt{n}}$$

With a t-distribution of df = 2, we compute a p-value of 0.0342067, confirming that our results are statistically significant at the 95% confidence level.

## 7 Conclusion

We examine why current LLMs fall short in real-world financial applications and verify this claim by creating a new benchmark, FinanceQA. Through testing current State-Of-The-Art models with FinanceQA, we conclude that LLMs have significant difficulty answering questions that mirror real-world financial analysis tasks, including both tactical and conceptual questions. Our findings highlight several insights about the current state of LLMs in finance. First, due to the sharp discontinuity in utility required in specialized fields, such as finance, these models provide minimal value to professionals who demand a high bar for precision and accuracy. Second, the complex nature of financial analysis, requiring both precise calculations and reasoning with incomplete information, demands a more comprehensive approach to model development and evaluation than current benchmarks provide. Our work suggests that meaningful progress in this domain will require training to prioritize high levels of financial rigor, compliance with accounting and valuation principles, and reasoning under incomplete information. Our paper is positioned to highlight the significant gap in professional requirements and current LLM capabilities while providing a framework for evaluating and developing models that meet the demands of real-world financial analysis.

## 8 Limitations

Our study was limited to evaluating the LLMs using solely tactical questions based on data from Costco, which may not generalize to other industries with different financial structures, such as healthcare or energy. Furthermore, financial professionals frequently work with Excel for financial modeling (e.g. DCF models and LBO models), and this area was not tested but is crucial for the practical adoption of LLMs in finance. Additionally, despite the improvements achieved through our synthetic data generation, it is likely that a full human-annotated dataset would have led to even higher accuracy. However, due to limited annotators, budget, and time, we were limited in the number of high-quality-curated examples.